\documentclass[lettersize,journal]{IEEEtran}
\usepackage{amsmath,amsfonts}
\usepackage{algorithmic}
\usepackage{algorithm}
\usepackage{array}
\usepackage[caption=false,font=normalsize,labelfont=sf,textfont=sf]{subfig}
\usepackage{textcomp}
\usepackage{stfloats}
\usepackage{url}
\usepackage{verbatim}
\usepackage{graphicx}
\usepackage{cite}
\usepackage{hyperref}
\usepackage{booktabs}
\begin{document}

\title{\textsc{ProtModernBERT}: Compute-Efficient Protein Language Modeling}

\author{
Alberto Martínez$^{1,\dagger}$,
Natalia Flechas Manrique$^{1,\dagger}$,
Llorenç Espinosa-Portalés$^{1, *}$,
and Luc Andrea$^{2}$
\thanks{* Corresponding author, e-mail: \href{mailto:llorenc.espinosa@multiversecomputing.com}{llorenc.espinosa@multiversecomputing.com}}
\thanks{$^{1}$ Multiverse Computing, Parque Científico y Tecnológico de Gipuzkoa,
Paseo de Miramón 170, 20014 Donostia-San Sebastián, Spain.}
\thanks{$^{2}$ Multiverse Computing, 7 rue de la Croix Martre, 91120 Palaiseau, Paris, France}
\thanks{$^{\dagger}$ These authors contributed equally to this work.}
}

\maketitle

\begin{abstract}
Large protein language models provide a powerful framework for learning sequence–function relationships and guiding experimental design. However, architectural improvements are underexplored, especially in low-compute regimes of relevance for resource-constrained research laboratories. We show that the training cost for encoder-only transformer architectures can be reduced by leveraging modern architectural refinements, offering an encouraging performance / compute-efficiency trade-off. We introduce \textsc{ProtModernBERT}, a modernized \textsc{BERT}-style model pre-trained on amino acid sequences, and compare it with \textsc{ProtBert}, \textsc{Amplify}, and \textsc{ESM2} across a set of supervised structural tasks — encompassing secondary structure prediction, disorder prediction, and subcellular localization — and ProteinGym’s zero-shot functional benchmark. We find that \textsc{ProtModernBERT} matches or closely tracks the performance of \textsc{Amplify-120M} and \textsc{ProtBert} with up to 80-90\% reduction in pre-training FLOPs, and narrows the gap to \textsc{ESM2} when controlling for pre-training. Our analysis using a cluster-based validation protocol for hyperparameter selection and randomly initialized models disentangles the contributions of architecture and data, showing that modern components provide stronger inductive biases that are more consequential than parameter count in low-compute regimes. 

\end{abstract}

\section{Introduction}

Transformer-based architectures \cite{vaswani2017attention, devlin2019bert} have been a crucial innovation for both predictive and generative modeling across multiple disciplines (e.g. \cite{dosovitskiy2021image,radford2021learning,jaegle2021perceiver}). The similarity between protein sequences and natural language —where amino acids function as a finite vocabulary forming higher-order functional structures \cite{asgari2015continuous,heinzinger2019modeling}— suggests that both domains are amenable to similar modeling techniques. Consequently, models such as \textsc{ProtBert} \cite{elnaggar2020protrans}, \textsc{ESM2} \cite{lin2023esm2}, and \textsc{Amplify} \cite{Fournier2024.09.23.614603} have successfully repurposed transformer designs for sequence-function prediction. We extend this trajectory by introducing ProtModernBERT, investigating whether recent architectural refinements can further reduce pre-training compute without degrading downstream performance.

Architectural innovations for language models have been the focus of intensive research aimed at improving both performance and efficiency. Notable examples include rotary positional embeddings as an alternative to absolute positional encodings \cite{su2021roformer}, improvements to the attention mechanism such as FlashAttention \cite{dao2022flashattention}, as well as revised normalization strategies and activation functions. While these advances have been critical for scaling modern LLMs, earlier architectures such as BERT \cite{devlin2019bert}, which remains widely used in industrial settings, have also been revisited and updated to incorporate these improvements, most notably in ModernBERT \cite{modernbert2024}. Importantly, recent work on genomic foundation models has suggested that downstream performance is not determined by pre-training scale alone, and that architectural choices can provide strong inductive biases even in randomly initialized models \cite{Vishniakov2024.12.18.628606}. In this spirit, we seek to update \textsc{ProtBert} \cite{elnaggar2020protrans} into a modernized variant, \textsc{ProtModernBERT}.

By introducing \textsc{ProtModernBERT}, which is a ModernBERT-derived encoder pre-trained on the UR100P corpus, we investigate whether recent architectural refinements can reduce pre-training compute for protein language models (PLMs) without degrading downstream performance. To this end, we compare it against \textsc{ProtBert}, \textsc{Amplify}, and \textsc{ESM2} under both pre-trained and randomly initialized (\emph{randinit}) conditions. These models span different architectural and training choices while remaining widely adopted benchmarks in the field. \textsc{ProtBert} represents the adaptation of the BERT-style masked language modeling paradigm to protein sequences; \textsc{ESM2} is a state-of-the-art transformer model extensively validated across diverse downstream biological tasks; and \textsc{Amplify} emphasizes training efficiency and modern architectural refinements. Together, they provide a meaningful basis for comparison in terms of architectural design, computational efficiency, and downstream performance. 

In our findings, we observe that across DeepLoc and secondary-structure benchmarks, \textsc{ProtModernBERT} matches or closely tracks the performance of \textsc{ProtBert} and \textsc{Amplify-120M} despite using substantially fewer pre-training FLOPs, and narrows the gap to \textsc{ESM2} once we control for pre-training. In addition, we adopt a more principled dataset partitioning strategy for pre-training, which reduces information leakage and better reflects realistic generalization scenarios. As part of this comparison, we also examine the role of rotary positional embeddings and other ModernBERT-style components in protein sequence modeling. As we will discuss later on, we do not observe consistent accuracy improvements from RoPE under our pre-training budget.

Finally, our experiments highlight that secondary structure prediction remains a challenging task for current PLMs. Despite architectural and training improvements, performance gains on secondary structure datasets are limited, suggesting that additional inductive biases or task-specific supervision may be necessary to better capture structural information. At the same time, tasks such as disorder prediction appear closer to saturation, with comparatively smaller gains from pre-training and modest separation between architectures, underscoring the task-dependent impact of both architecture and pre-training compute.

\section{Methods}

This section is organized into three parts. In Section~\ref{subsection:Model pre-training}, we describe the tokenizer, pre-training dataset, architectural choices, and optimization hyperparameters used to train \textsc{ProtModernBERT}, and summarize the final pre-training run. In Section~\ref{subsection: model fine-tuning and Evaluation}, we present the fine-tuning and evaluation protocols for both pre-trained and randomly initialized models across DeepLoc and secondary-structure benchmarks, and detail how we compare \textsc{ProtModernBERT} against \textsc{ProtBert}, \textsc{Amplify}, and \textsc{ESM2}. Finally, in Section~\ref{sec:hardware_specifications}, we report the hardware configuration used for pre-training and fine-tuning.
\subsection{Model Pre-Training}
\label{subsection:Model pre-training}

\subsubsection{Tokenizer}

We use a tokenizer that assigns one token per amino acid, following \textsc{ProtBert} \cite{elnaggar2020protrans}, ESM2 and \textsc{Amplify}. The vocabulary contains 30 tokens: the 20 canonical amino acids, the special tokens [PAD], [UNK], [CLS], [SEP], and [MASK], following BERT \cite{devlin2019bert}, and the additional residue symbols ‘X’, ‘U’, ‘B’, ‘Z’, and ‘O’. Although these non-canonical amino acids are not present in the pre-training data, these tokens are retained to ensure compatibility with downstream datasets, where ambiguous or rare residues (e.g., ‘X’ or selenocysteine ‘U’) may occur, avoiding the need for vocabulary changes during fine-tuning.

We adopt a one-token-per-amino-acid representation. This approach provides explicit control over sequence length for design tasks, facilitates the learning of residue-level embeddings aligned with physicochemical properties, and remains computationally feasible given the manageable length of most protein sequences.

\subsubsection{Pre-Training Dataset}

For \textsc{ProtModernBERT} we adopt the \textsc{UR100P} dataset, described in \cite{Fournier2024.09.23.614603} and originally used to train \textsc{Amplify}. It aggregates sequences from UniProt/UniRef100\cite{uniprot2023uniprot}, the Observed Antibody Space (OAS) \cite{kovaltsuk2018observed}, and SCOP 2 \cite{chandonia2022scop2} (December 2023 release). Unlike aggressively clustered corpora (e.g., UniRef50), \textsc{UR100P} preserves naturally occurring sequence redundancy, treating repeated observations as a signal of biological confidence while reducing the relative weight of singleton or low-quality sequences. Sequences shorter than five residues or containing ambiguous or non-canonical amino acids were excluded. 

To ensure consistency with existing efficiency benchmarks, we utilize the UniProt validation set established by \textsc{Amplify} \cite{Fournier2024.09.23.614603} rather than constructing a new reference corpus. This validation data was designed to support targeted evaluation across distinct regions of protein sequence space and consists of three subsets: (i) 10,000 reference proteome sequences from high-quality assemblies (BUSCO completeness $>80\%$) spanning bacteria, archaea, and eukaryotes; (ii) 10,000 paired OAS antibody heavy–light chain sequences clustered at 90\% sequence identity; and (iii) 10,000 SCOP family-representative domain sequences clustered at 30\% identity.

Next, training sequences were drawn from the remaining UniProt, UniRef, SCOP, and OAS data after removing any sequence with greater than 90\% identity to validation entries. Paired antibody sequences were included in both heavy$|$light and light$|$heavy orientations. The resulting training corpus contains approximately 364M sequences and provides broad coverage while minimizing information leakage into validation sets.

This dataset choice for \textsc{ProtModernBERT} sets a clear difference with \textsc{ProtBert}, originally pre-trained directly on UniRef100 \cite{10.1093/bioinformatics/btm098}. In contrast, we pre-trained \textsc{ProtModernBERT} on \textsc{UR100P}, motivated both by the considerations discussed above and by empirical evidence reported \cite{Fournier2024.09.23.614603}, showing that models trained on UniRef100 augmented with OAS antibody sequences and SCOP domains (\textsc{UR100P}) exhibit faster learning dynamics than those trained on UniRef50 (used by e.g. ESM2 and ProtGPT2), likely due to the increased scale and diversity of the training corpus.

\subsubsection{Model Implementations and Architectural Context}

As previously stated, the primary objective of this work is to train and evaluate \textsc{ProtModernBERT}, a protein language model derived from the \textsc{ModernBERT} architecture. Existing models were studied throughout this work with two distinct roles: (i) guiding architectural and hyperparameter decisions during pre-training, and (ii) providing representative baselines for downstream evaluation.

To guide architectural design and hyperparameter selection, we focused on \textsc{ModernBERT} and \textsc{Amplify}. ModernBERT serves as the direct architectural precursor to \textsc{ProtModernBERT}, while \textsc{Amplify} represents the closest existing protein-specific BERT-style model trained on the same underlying dataset (\textsc{UR100P}). Together, these models provide a principled reference for adapting a modern language architecture to the protein domain.

For downstream evaluation, we additionally included \textsc{ProtBert} and \textsc{ESM2}. \textsc{ProtBert} represents an earlier generation of protein BERT models trained on UniRef, while \textsc{ESM2} is a strong, widely adopted baseline trained on a large and diverse protein corpus. These models were included to contextualize performance rather than to inform architectural or optimization choices.

All models were evaluated using publicly released, reference implementations to ensure reproducibility and to avoid confounding effects arising from implementation-specific optimizations. Specifically, we used the official HuggingFace \cite{wolf2020transformers} implementations of \textsc{ProtBert}, \textsc{Amplify}, \textsc{ModernBERT}, and \textsc{ESM2}.

\textsc{ProtModernBERT} integrates modern transformer elements detailed in Table \ref{tab:hyperparams}, including bias-free linear layers, FlashAttention, and Rotary Positional Embeddings (RoPE). While we follow the \textsc{ModernBERT} template for depth, hidden size, and local attention windows, we adopt \textsc{Amplify}’s higher weight decay (0.01) and include mild attention dropout (p=0.1) to stabilize training on the repetitive UR100P corpus. Detailed parameter comparisons are provided in Table \ref{tab:hyperparams}.

\begin{table*}[t]
\centering
\small
\caption{Key architectural and optimization hyperparameters for \textsc{ModernBERT} (Base), \textsc{Amplify} (Base), and \textsc{ProtModernBERT}. These models are shown because they directly inform the architectural design and hyperparameter selection of \textsc{ProtModernBERT}.}
\label{tab:hyperparams}
\resizebox{\textwidth}{!}{
\vspace{-1em} 
\begin{tabular}{lccc}
\toprule
\textbf{Hyperparameter} & \textbf{ModernBERT (Base, 140M)} & \textbf{\textsc{Amplify}  (Base, 120M)} &\textbf{ProtModernBERT} \\
\midrule
Optimizer & StableAdamW ($\beta{=}(0.90, 0.98)$, $\epsilon{=}1\text{e-}6$) & AdamW ($\beta{=}(0.90, 0.95)$, $\epsilon{=}1\text{e-}8$) & AdamW ($\beta{=}(0.90, 0.99)$, $\epsilon{=}1\text{e-}8$)\\
Learning rate & $8\text{e-}4$ (pretrain), $3\text{e-}4$ (context) & 0.001 & $8\text{e-}4$ \\
Weight decay & $1\text{e-}5$ & 0.01 & 0.01\\
LR schedule & Trapezoidal & Cosine decay (Stage 1), none (Stage 2) & Trapezoidal \\
Global Batch size & 4,608 (pretrain), 72 (context) & 4,096 & 12288\\
Precision & fp16/bf16 (via FlashAttention) & bf16 (mixed precision) & bf16 (via FlashAttention) \\
Gradient clipping & – & 1.0 & - \\
Training strategy & DDP & DeepSpeed ZeRO Stage~3 & DDP \\
Hardware & 8$\times$H100 GPUs & – (DeepSpeed setup) & 96$\times$A100 GPUs\\
Vocabulary size & 50,368 & 27 & 30\\
Unused tokens & 83 & – & -\\
Layers & 22 & 24 & 22\\
Hidden size & 768 & 640 & 768\\
Intermediate (FFN) size & 1,152 & 2,560 & 1152\\
GLU expansion & 2,304 & – & -\\
Attention heads & 12 & 10 & 12 \\
Activation & GeGLU & SwiGLU & GeGLU\\
Norm $\epsilon$ & $1\text{e-}5$ & $1\text{e-}5$ & $1\text{e-}5$ \\
Dropout (attn out) & 0.1 & 0 & 0.1\\
Dropout (other) & 0.0 & – & -\\
Max seq. length & 1,024 (pretrain), 8,192 (context) & 512 (Stage~1), 2,048 (Stage~2) & 1,024 (pretrain) \\
RoPE $\theta$ & 160{,}000 (local: 10{,}000) & 10{,}000 & 160{,}000 (local: 10{,}000) \\
Local attn. window & 128 & – & 128\\
Global attention & Every 3 layers & – & Every 3 layers\\
Sequences & – & 364M & 364M \\
Tokens & 2T & 137B & 137B \\
\bottomrule
\end{tabular}
}
\vspace{0.5em} 
\vspace{-1em} 
\end{table*}

\subsubsection{Hyperparameter Selection}

To inform the final hyperparameter configuration of \textsc{ProtModernBERT}, we conducted four short preliminary experiments focusing on the behavior of batch size and weight decay during pre-training, analyzing training stability and loss dynamics. These experiments were designed to complement architectural guidance from closely related models. In particular, we compared against \textsc{Amplify}, which shares the same pre-training dataset \textsc{UR100P}, and \textsc{ModernBERT}, which serves as the direct architectural basis for our model. Leveraging these two reference points, we explored which optimization settings are best suited for a modern BERT-style protein language model trained at moderate compute scale.

Preliminary experiments (see Appendix ~\ref{app:preliminary}) informed our optimization strategy. We selected a weight decay of 0.01 to ensure training stability and a trapezoidal learning rate schedule for its checkpoint continuation benefits. A local batch size of 128 was used to maximize hardware utilization without compromising generalization.

\subsubsection{Final Pre-Training Run}

Based on the preliminary experiments, we selected the best-performing checkpoint and hyperparameter configuration to carry out a longer pre-training run (see \ref{app:preliminary} for further information). This final experiment was trained for 20 hours using 96 NVIDIA A100 GPUs. During this run, \textsc{ProtModernBERT} achieved an average hardware utilization efficiency of 48\%, measured as the ratio between achieved and theoretical peak FLOPs. The model exhibits stable training dynamics and consistent loss reduction throughout the run. Specifically, \textsc{ProtModernBERT} was trained for 10,887 gradient steps with a global batch size of 12,288 sequences (1024 context), totaling 137B tokens

\subsection{Model Fine-tuning and Evaluation}
\label{subsection: model fine-tuning and Evaluation}



\subsubsection{Fine-Tuning Datasets}

In this section, we describe the datasets that were used to train, validate and evaluate our models. Table~\ref{tab:datasets} summarizes dataset sizes and labels.

\begin{table}
\vskip 0.15in
\centering
\small
\caption{Overview of benchmark datasets.}
\label{tab:datasets}
\begin{tabular}{lccc}
\toprule
\textbf{Dataset} & \textbf{Task} & \textbf{Train} & \textbf{Test} \\
\midrule
DeepLoc & Localization (10-class) & 6,622 & 1,842 \\
        & Membrane (binary) & 6,622 & 1,842 \\
\midrule
Train\_HHblits & Sec. structure & 10,792 & -- \\
CASP12 & Sec. structure & -- & 20 \\
TS115  & Sec. structure & -- & 115 \\
CB513  & Sec. structure & -- & 511 \\
NEW364 & Sec. structure & -- & 364 \\
\bottomrule
\end{tabular}
\end{table}

We evaluate performance across three standard benchmarks summarized in Table \ref{tab:datasets}: DeepLoc \cite{almagro2017deeploc} for 10-class subcellular localization and binary membrane classification; the HHblits-based suite - CASP12\cite{abriata2018casp12}, TS115\cite{cuff1999evaluation}, CB513\cite{yang2018sixtyfive} - for 3-state and 8-state secondary structure and disorder prediction; and NEW364 \cite{elnaggar2020protrans} to assess generalization on non-redundant proteins released after 2019.

\subsubsection{Fine-Tuning Experiments}
\label{subsubsection:Fine-Tuning Experiments}

We conduct two families of fine-tuning experiments: one using pre-trained checkpoints and one starting from randomly initialized parameters (\emph{randinit}). The randinit setting allows us both to isolate the contribution of pre-training for \textsc{ProtModernBERT} and to compare architectures on a more equal footing, without confounding differences in pre-training data or FLOPs. Downstream performance across both pre-trained and randomly initialized (randinit) conditions is summarized in Figure ~\ref{fig:results_accuracy} (Accuracy) and Figure ~\ref{fig:results_f1} (Macro F1). We will further comment these results in Section \ref{sec:results_discussion}

For each model–task pair and initialization condition, we perform a grid search over learning rate \([3\text{e-}5, 5\text{e-}5]\), warmup epochs \([4, 6]\), and weight decay \([0.02, 0.1]\). All runs use batch size \(=16\), warmup rate \(=0.05\), and a maximum sequence length of 512 tokens. More information about the final choice of parameters can be found in the Appendix (\ref{app:Fine-tuning results}).

\subsubsection{Functional validation: ProteinGym}
\label{subsubsection:Functional validation: ProteinGym}
While ProteinGym \cite{ProteinGym} is a comprehensive benchmark encompassing a wide range of tasks—including clinical variant prioritization and indel effect prediction—we focus our evaluation on the zero-shot substitution benchmark. This task specifically assesses a model's ability to capture sequence–function relationships without further task-specific supervision. We measure the model's performance by calculating the Spearman correlation between the model-assigned log-likelihoods—specifically the likelihood of the mutated sequence relative to the wild-type—and experimental fitness scores derived from deep mutational scanning (DMS) assays. This targeted evaluation provides a direct measure of the model's functional "semantics" across diverse categories such as organismal fitness, binding, and protein stability.

\subsection{Hardware Specifications}
\label{sec:hardware_specifications}

During pre-training we used the following hardware specifications: 12 eight-GPU accelerated compute nodes, each with 2 AMD Milan EPYC 7543 processors (32 cores at 2.80 GHz, totaling 64 cores per node), 512 GB of memory, and 8 Nvidia A100 SXM4 80 GB GPUs.  For fine-tuning, we used only one NVIDIA A10G Tensor Core.


\begin{figure*}[t]
\centering
\includegraphics[width=0.95\textwidth]{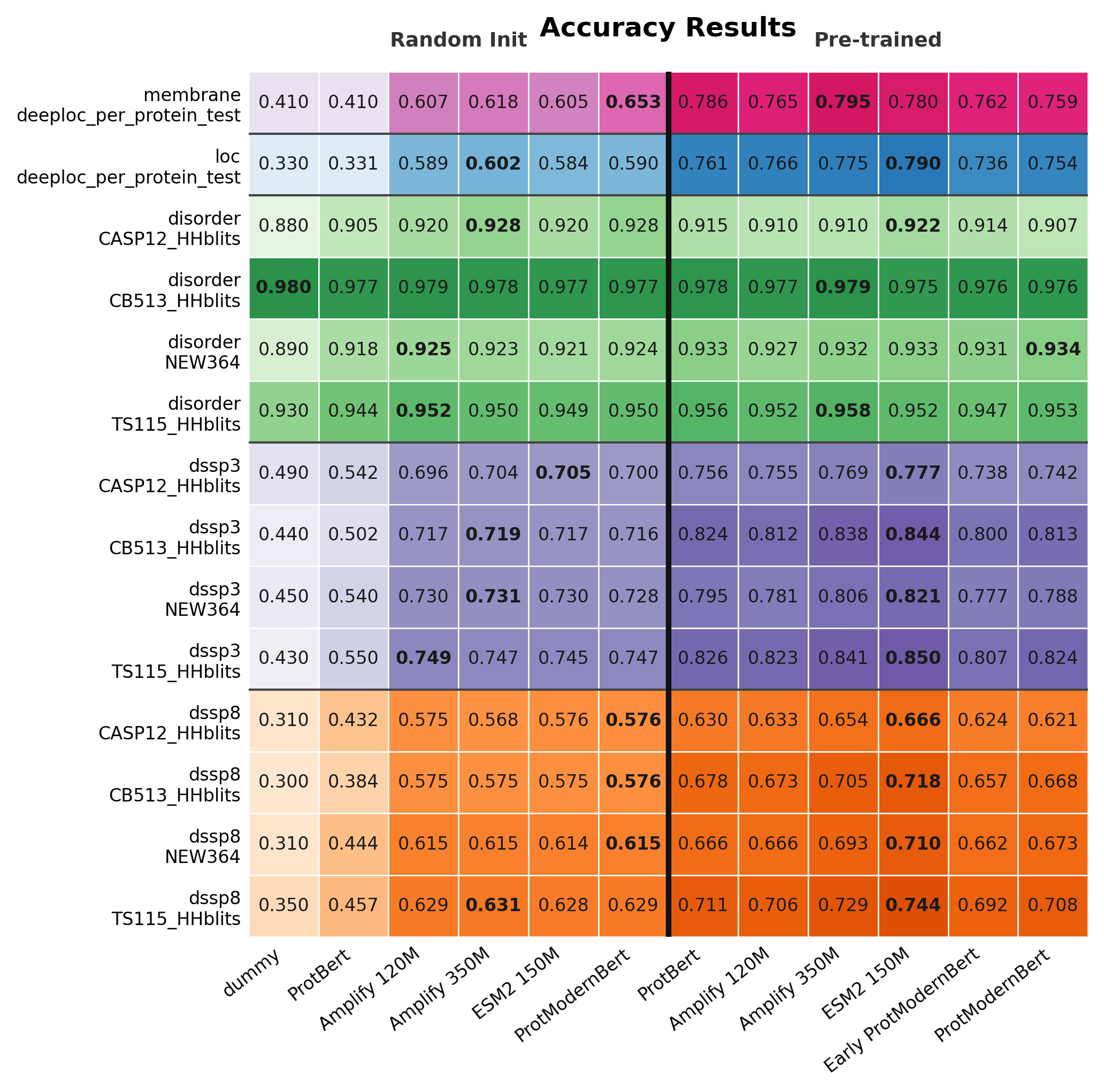}
\caption{Accuracy benchmark results under the \textbf{pre-trained} and \textbf{random initialization} conditions for \textsc{ProtBert}, \textsc{Amplify-120M}, \textsc{Amplify-350M}, and \textsc{ProtModernBERT}. The dummy majority-class baseline is shown on the far left for scale.}
\label{fig:results_accuracy}
\end{figure*}

\begin{figure*}[t]
\centering
\includegraphics[width=0.95\textwidth]{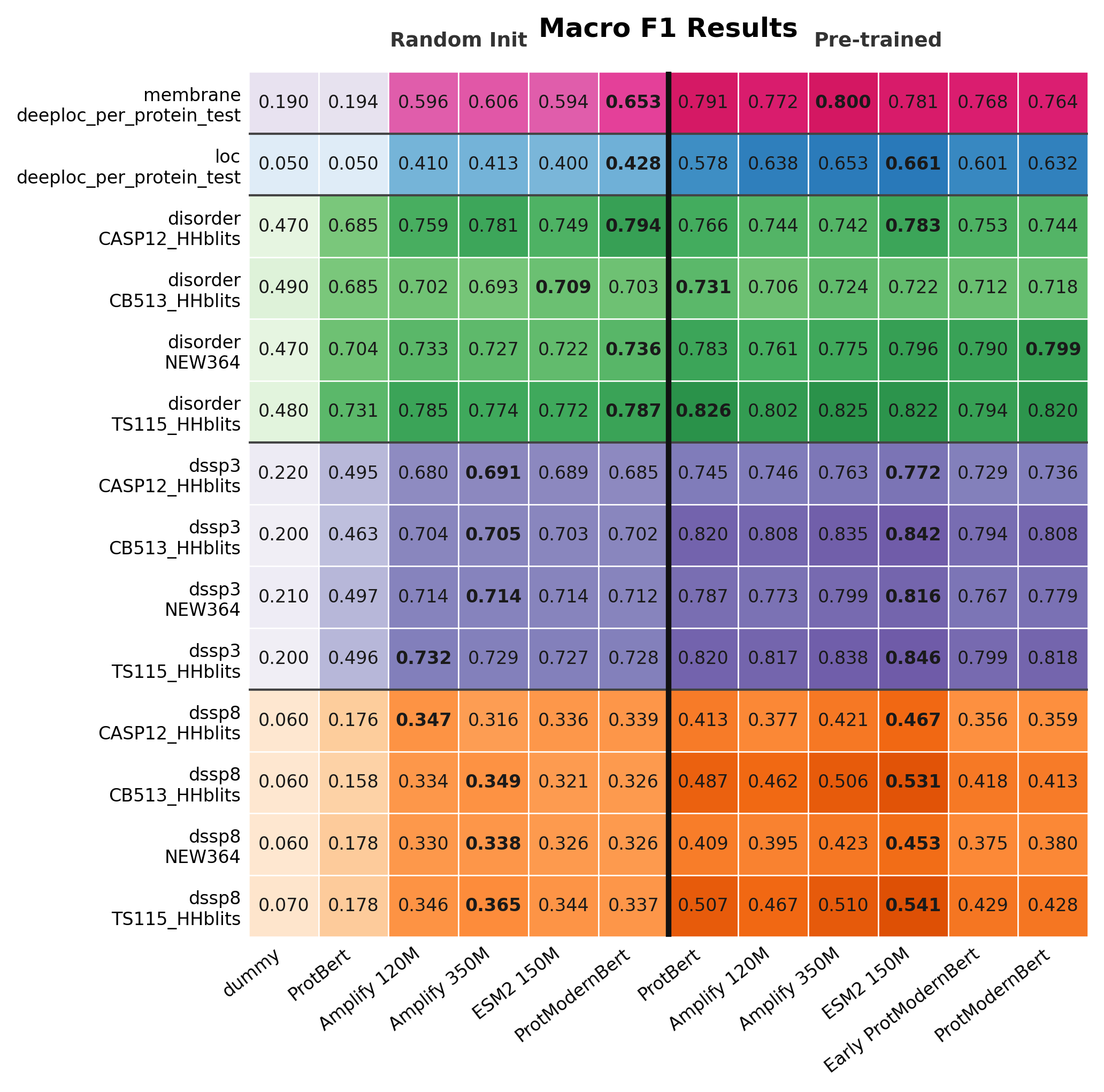}
\caption{F1 benchmark results under the \textbf{pre-trained} and \textbf{random initialization} conditions for \textsc{ProtBert}, \textsc{Amplify-120M}, \textsc{Amplify-350M}, and \textsc{ProtModernBERT}. As in the previous figure, the dummy majority-class baseline is shown on the far left for scale.}
\label{fig:results_f1}
\end{figure*}


\section{Results and Discussion}
\label{sec:results_discussion}
\renewcommand{\thefootnote}{\fnsymbol{footnote}}

We evaluate \textsc{ProtModernBERT} against \textsc{AMPLIFY-120M}, \textsc{AMPLIFY}-350M, \textsc{ESM2-150M}, and \textsc{ProtBert} 420M. In addition, we report results from an earlier \textsc{ProtModernBERT} checkpoint trained to assess early-stage learning behavior. All models are evaluated under both pre-trained and randomly initialized (\emph{randinit}) conditions. Table  \ref{tab:model_comparison} shows a model comparison in terms of parameters and FLOPs.

Overall, our results indicate that \textsc{ProtModernBERT} is competitive with larger and more computationally expensive baselines, despite requiring substantially less pre-training compute. This suggests that modern architectural choices can yield strong representations even under constrained training regimes.

\begin{table}[H]
\centering
\small
\caption{Comparison of model size and pre-training compute.}
\label{tab:model_comparison}
\begin{tabular}{lccc}
\toprule
\textbf{Model} & \textbf{Parameters} & \textbf{FLOPs}  \\
\midrule
\textsc{ProtBert} & 420M & $6.0\times10^{21}$  \\
\textsc{ESM2-150M} & 150M & $9\times10^{20}$ \footnotemark   \\
\textsc{Amplify-120M} & 120M & $1.6\times10^{21}$  \\
\textsc{Amplify-350M} & 350M & $4.6\times10^{21}$ \\
\textsc{ProtModernBERT} & 140M & $5.8\times10^{20}$ \\
Early \textsc{ProtModernBERT} & 140M & $1.3\times10^{20}$ \\
\bottomrule
\end{tabular}
\end{table}
\footnotetext{Training FLOPs for \textsc{ESM2-150M} are not reported in original ESM2 publications.
\textsc{ESM2-150M} does not seem to be the most computationally expensive model in our comparison:
based on our estimates, its total pre-training budget lies in the range
$9\times10^{20}$--$1\times10^{21}$ FLOPs, we report the most conservative estimate and further discuss this in the Appendix (\ref{app:esm2-compute}).}

\paragraph{Compute Efficiency and Aggregate Performance}
Across benchmarks, \textsc{ProtModernBERT} consistently matches or closely tracks the performance of \textsc{Amplify-120M} and \textsc{ProtBert} while using the smallest pre-training budget among the evaluated models.  
Within the \textsc{AMPLIFY} family, the 350M model offers only modest gains over the 120M variant, even under pre-training, whereas the difference between pre-trained and \emph{randinit} variants is much larger. 
 \textsc{ESM2} typically achieves the strongest absolute performance, despite being trained with a total pre-training FLOP budget that is lower than that of \textsc{Amplify-120M} and \textsc{ProtBert}. This efficiency is particularly notable given that \textsc{ProtBert} and \textsc{Amplify-350M} benefit from roughly an order of magnitude more compute, yet \textsc{ESM2} often performs best. We hypothesize that this reflects advantages in corpus design and curation, masking strategy and training setup, rather than simply having the largest model or the largest compute budget.
  Interestingly, \textsc{ESM2} advantage over \textsc{ProtModernBERT} narrows considerably once we control for pre-training (i.e., in the \emph{randinit} setting).
  
Together, these observations suggest that, in the low- to moderate-compute regime, architectural refinements and pre-training quality are more consequential than further increases in parameter count or training budget.

\paragraph{Task-Dependent Impact of Pre-Training}
The benefit of pre-training is strongly task dependent. For multi-class, structurally demanding tasks such as secondary structure prediction and subcellular localization, the gap between pre-trained and randinit models is substantial in both accuracy and macro F1. By contrast, for disorder prediction, random-initialized models already achieve strong performance and pre-training yields comparatively smaller gains, particularly in accuracy. This pattern suggests that pre-training is most valuable for tasks that require richer, long-range sequence features or that rely on underrepresented classes, whereas simpler or more locally determined tasks may be closer to saturation.
For all datasets, the dummy majority-class baseline (integrated into the left-most column of Figures ~\ref{fig:results_accuracy}  and ~\ref{fig:results_f1}) performs as expected: it attains moderate accuracy on highly imbalanced tasks like disorder prediction but fails to provide meaningful macro-averaged F1 in multi-class settings. This highlights that non-trivial sequence information is required to achieve our reported downstream results, and that improvements over this baseline reflect genuine gains beyond class imbalance effects. Appendix ~\ref{app:dummy-baseline} further explains how the dummy baseline is calculated.

\paragraph{Disorder Prediction}
On the \textbf{Disorder} benchmarks, differences across models are modest: all pre-trained models reach high accuracy and macro F1. Accuracy changes little between pre-trained and randinit settings, but macro F1 improves with pre-training across architectures, suggesting better calibration and class balance. The relative gain is smaller for \textsc{ProtModernBERT}, whose randinit variant is already competitive, consistent with useful inductive biases for this largely local task.

\paragraph{Secondary Structure Prediction (DSSP3 / DSSP8)}
Unlike disorder prediction, \textbf{DSSP3}—and especially \textbf{DSSP8}—remain challenging with clear headroom. When pre-trained, \textsc{ESM2} performs best, underscoring the value of large-scale pre-training for structurally demanding tasks. \textsc{ProtBert} is competitive when pre-trained but collapses under randinit, with macro F1 near chance, indicating limited inductive bias for secondary structure without pre-training.

By comparison, \textsc{ProtModernBERT}, \textsc{Amplify}, and \textsc{ESM2} generalize substantially better in the randinit setting, maintaining non-trivial accuracy and macro F1 across CASP12, CB513, NEW364, and TS115. Notably, without pre-training the gap between \textsc{ESM2} and \textsc{ProtModernBERT} largely disappears, pointing to architectural comparability when data scale and compute are controlled. \textsc{Amplify-120M} and \textsc{Amplify-350M} perform similarly on DSSP3/8 despite the parameter gap, suggesting diminishing returns from added size within this family for secondary structure tasks.

\paragraph{Subcellular Localization (DeepLoc)}
On the \textbf{DeepLoc} benchmarks, \textsc{ProtBert} degrades sharply under randinit—especially in macro F1—while its pre-trained variant remains strong. This sensitivity may reflect reliance on absolute positional embeddings and earlier normalization choices that provide weaker inductive bias for learning global sequence properties from scratch. In contrast, architectures with rotary embeddings and more modern attention/normalization (\textsc{Amplify}, \textsc{ESM2}, and \textsc{ProtModernBERT}) retain substantially better randinit performance, highlighting stronger architectural priors.

In the pre-trained setting, \textsc{ESM2} and \textsc{Amplify-350M} achieve the best overall DeepLoc accuracy and macro F1. \textsc{ProtModernBERT} remains competitive but underperforms on some subtasks, particularly membrane classification where \textsc{Amplify} and \textsc{ESM2} achieve higher macro F1. Fully closing this gap may require more pre-training compute, more targeted pre-training data, or task-specific adaptation (for example, multi-task training or localization-aware objectives).

\paragraph{Mutational Effect Prediction on ProteinGym}

In line with our functional validation protocol, we report the results for the ProteinGym zero-shot substitution task. As shown in Table~\ref{tab:proteingym_results}, \textsc{ProtModernBERT} (0.373) matches the functional predictive power of \textsc{ProtBert} (0.376), while trailing the more computationally expensive ESM2-150M (0.387) baseline on the aggregate benchmark. 

The results highlight that despite ESM2’s advantage—driven by its \textbf{7.5$\times$ larger pre-training data volume} (1.024T vs. 137B tokens) —\textsc{ProtModernBERT} remains highly efficient. Notably, it outperforms ESM2 on Organismal Fitness (0.379 vs. 0.305). The measurable gap observed for the Early \textsc{ProtModernBERT} checkpoint (0.327) further indicates that the full pre-training duration was necessary to capture these complex functional signals. These findings reinforce the conclusion that architectural modernization enables the model to effectively learn functional landscapes with 80--90\% fewer pre-training FLOPs.

\begin{table*}[t]
\centering
\small
\caption{ProteinGym zero-shot substitution results (Spearman correlation). \textsc{ProtModernBERT} demonstrates competitive functional modeling across categories, notably achieving the highest score in Organismal Fitness despite significantly lower pre-training compute.}
\label{tab:proteingym_results}
\begin{tabular}{lcccccc}
\toprule
\textbf{Model} & \textbf{Total (Overall)} & \textbf{Org. Fitness} & \textbf{Binding} & \textbf{Expression} & \textbf{Activity} & \textbf{Stability} \\
\midrule
\textsc{ProtBert} & 0.376 & 0.357 & 0.211 & 0.338 & 0.363 & 0.448 \\
\textsc{ProtModernBERT} & 0.373 & \textbf{0.379} & 0.282 & 0.385 & 0.342 & 0.400 \\
Early \textsc{ProtModernBERT} & 0.327 & 0.331 & 0.288 & 0.354 & 0.316 & 0.331 \\
ESM2-150M & \textbf{0.387} & 0.305 & \textbf{0.326} & \textbf{0.402} & \textbf{0.391} & \textbf{0.510} \\
\bottomrule
\end{tabular}
\end{table*}

\paragraph{Effect of rotary positional embeddings.}
Because RoPE has been proposed as a mechanism to better capture long-range dependencies \cite{su2021roformer}, we examined its impact in the context of \textsc{ProtModernBERT}. Within our experimental setting and pre-training budget, we do not observe consistent accuracy/F1 gains from RoPE compared to architectures relying on absolute positional encodings. Instead, its benefits appear to manifest primarily in terms of training stability and computational efficiency, particularly when combined with modern attention implementations such as FlashAttention. We emphasize that this should be interpreted as an empirical observation under a specific training regime and moderate compute budgets rather than a definitive assessment of RoPE for protein sequences, as we did not conduct exhaustive ablation experiments over positional encoding schemes or training budgets.

\section{Conclusion}

We present \textsc{ProtModernBERT}, a compute-efficient protein language model that matches the performance of benchmarks like \textsc{Amplify-120M} and \textsc{ProtBert} while reducing pre-training compute by 80--90\%. Our results demonstrate that modern architectural components enable custom model training to be feasible even for resource-constrained laboratories. 

By jointly analyzing pre-trained and randinit settings, we disentangle the contributions of architecture, data, and compute. Across structural tasks, pre-training consistently yields the largest gains, yet \textsc{ProtModernBERT} exhibits markedly better performance than \textsc{ProtBert} under random initialization. This suggests that modern architectural refinements provide stronger inductive biases for structural prediction, narrowing the performance gap to state-of-the-art models like \textsc{ESM2} once data scale and compute are controlled.

Evaluation on the \textsc{ProteinGym} zero-shot benchmark confirms that \textsc{ProtModernBERT} effectively captures functional landscapes, notably outperforming \textsc{ESM2} on the Organismal Fitness subtask despite using an order of magnitude less compute. These findings indicate that principled architectural design—incorporating modern attention and normalization—can drive progress even under realistic data constraints.

\section{Limitations}
Since comparisons occur at early pre-training stages, relative rankings might shift under longer training schedules. Differences in tokenization and optimization choices influence the attribution of gains, and the current lack of variance estimates limits the assessment of statistical significance. Additionally, some benchmarks like disorder prediction appear closer to saturation, while observations regarding specific components like \textsc{RoPE} are restricted to our current compute regime. \textsc{ProtModernBERT}’s reliance on the redundant UR100P corpus may bias the model toward a ``consensus grammar,'' potentially limiting reliability for rare or evolutionarily divergent proteins. Finally, we acknowledge the methodological challenge of fully disentangling architectural gains from the specific influence of pre-training data quality.

\section*{Data access}

The data used in this study is available online under the following license terms: UR100P, under CC-BY-4.0; ProteinGym, under MIT license; the fine-tuning datasets (DeepLoc, CASP12, TS115, CB513, and NEW364), under MIT license.  The models used in this study for benchmarking against \textsc{ProtModernBERT} (\textsc{ProtBert}, \textsc{ESM2}, and \textsc{Amplify}) are released under MIT license. The training and evaluation code and trained model weights for \textsc{ProtModernBERT} are not publicly available at this time.

\section*{Acknowledgements}

This work was granted access to the HPC resources of IDRIS under the allocation 2024-A0171015266  made by GENCI.



\bibliographystyle{IEEEtran}
\bibliography{bibliography}

@article{almagro2017deeploc,
  title={DeepLoc: prediction of protein subcellular localization using deep learning},
  author={Almagro Armenteros, Jose Juan and Sønderby, Casper Kaae and Sønderby, Søren Kaae and Nielsen, Henrik and Winther, Ole},
  journal={Bioinformatics},
  volume={33},
  number={21},
  pages={3387--3395},
  year={2017},
  publisher={Oxford University Press}
}

@article{cb513,
  title={Assessment of the CASP4 predictions for new and old targets},
  author={Cuff, James A and Barton, Geoffrey J},
  journal={Proteins: Structure, Function, and Genetics},
  volume={45},
  pages={38--46},
  year={2001},
  publisher={Wiley Online Library}
}

@article{kovaltsuk2018observed,
  title={Observed Antibody Space: a resource for data mining next-generation sequencing of antibody repertoires},
  author={Kovaltsuk, Andrei and Leem, Joseph and Kelm, Sebastian and Snowden, Jonathan and Deane, Charlotte M},
  journal={Journal of Immunology},
  volume={201},
  number={8},
  pages={2502--2509},
  year={2018},
  publisher={American Association of Immunologists}
}

@article{chandonia2022scop2,
    author = {Chandonia, John-Marc and Guan, Lindsey and Lin, Shiangyi and Yu, Changhua and Fox, Naomi K and Brenner, Steven E},
    title = {SCOPe: improvements to the structural classification of proteins – extended database to facilitate variant interpretation and machine learning},
    journal = {Nucleic Acids Research},
    volume = {50},
    number = {D1},
    pages = {D553-D559},
    year = {2022},
    month = {01},
    doi = {10.1093/nar/gkab1054},
    url = {https://doi.org/10.1093/nar/gkab1054},
    eprint = {https://academic.oup.com/nar/article-pdf/50/D1/D553/42058119/gkab1054.pdf},
}

@article{10.1093/bioinformatics/btm098,
    author = {Suzek, Baris E. and Huang, Hongzhan and McGarvey, Peter and Mazumder, Raja and Wu, Cathy H.},
    title = {UniRef: comprehensive and non-redundant UniProt reference clusters},
    journal = {Bioinformatics},
    volume = {23},
    number = {10},
    pages = {1282-1288},
    year = {2007},
    month = {03},
    issn = {1367-4803},
}

@article{vaswani2017attention,
  title={Attention Is All You Need},
  author={Vaswani, Ashish and Shazeer, Noam and Parmar, Niki and others},
  journal={Advances in Neural Information Processing Systems},
  year={2017}
}

@article{asgari2015continuous,
  title={Continuous Distributed Representation of Biological Sequences for Deep Proteomics and Genomics},
  author={Asgari, Ehsaneddin and Mofrad, Mohammad R.K.},
  journal={PLOS ONE},
  year={2015}
}

@article{heinzinger2019modeling,
  title={Modeling aspects of the language of life through transfer‐learning protein sequences},
  author={Heinzinger, Michael and Elnaggar, Ahmed and Wang, Yu and Dallago, Christian and Nechaev, Dmitrii and Matthes, Florian and Rost, Burkhard},
  journal={BMC Bioinformatics},
  volume={20},
  number={1},
  pages={723},
  year={2019},
  doi={10.1186/s12859-019-3220-8},
  url={https://doi.org/10.1186/s12859-019-3220-8}
}

@article{elnaggar2020protrans,
  author={Elnaggar, Ahmed and Heinzinger, Michael and Dallago, Christian and Rehawi, Ghalia and Wang, Yu and Jones, Llion and Gibbs, Tom and Feher, Tamas and Angerer, Christoph and Steinegger, Martin and Bhowmik, Debsindhu and Rost, Burkhard},
  journal={IEEE Transactions on Pattern Analysis and Machine Intelligence}, 
  title={ProtTrans: Toward Understanding the Language of Life Through Self-Supervised Learning}, 
  year={2022},
  volume={44},
  number={10},
  pages={7112-7127},
  doi={10.1109/TPAMI.2021.3095381}
}

@article {Fournier2024.09.23.614603,
	author = {Fournier, Quentin and Vernon, Robert M. and van der Sloot, Almer and Schulz, Benjamin and Chandar, Sarath and Langmead, Christopher James},
	title = {Protein Language Models: Is Scaling Necessary?},
	elocation-id = {2024.09.23.614603},
	year = {2024},
	doi = {10.1101/2024.09.23.614603},
	publisher = {Cold Spring Harbor Laboratory},
	URL = {https://www.biorxiv.org/content/early/2024/09/23/2024.09.23.614603},
	eprint = {https://www.biorxiv.org/content/early/2024/09/23/2024.09.23.614603.full.pdf},
	journal = {bioRxiv}
}

@article{su2021roformer,
  title={RoFormer: Enhanced Transformer with Rotary Position Embedding},
  author={Su, Jianlin and Lu, Yu and Pan, Shengfeng and Wen, Bo and Liu, Yunfeng},
  journal={arXiv preprint arXiv:2104.09864},
  year={2021}
}

@article{dao2022flashattention,
  title={FlashAttention: Fast and Memory-Efficient Exact Attention with IO-Awareness},
  author={Dao, Tri and others},
  journal={Advances in Neural Information Processing Systems},
  year={2022}
}

@article{modernbert2024,
      title={Smarter, Better, Faster, Longer: A Modern Bidirectional Encoder for Fast, Memory Efficient, and Long Context Finetuning and Inference}, 
      author={Benjamin Warner and Antoine Chaffin and Benjamin Clavié and Orion Weller and Oskar Hallström and Said Taghadouini and Alexis Gallagher and Raja Biswas and Faisal Ladhak and Tom Aarsen and Nathan Cooper and Griffin Adams and Jeremy Howard and Iacopo Poli},
      year={2024},
      eprint={2412.13663},
      archivePrefix={arXiv},
      primaryClass={cs.CL},
      url={https://arxiv.org/abs/2412.13663}, 
}

@article{lin2023esm2,
  title={Evolutionary-scale Prediction of Atomic-Level Protein Structure with a Language Model},
  author={Lin, Zeming and others},
  journal={Science},
  year={2023}
}

@article{uniprot2023uniprot,
  title={UniProt: the universal protein knowledgebase in 2023},
  author ={The UniProt Consortium},
  journal={Nucleic acids research},
  volume={51},
  number={D1},
  pages={D523--D531},
  year={2023},
  publisher={Oxford University Press}
}

@inproceedings{wolf2020transformers,
  title={Transformers: State-of-the-art natural language processing},
  author={Wolf, Thomas and Debut, Lysandre and Sanh, Victor and Chaumond, Julien and Delangue, Clement and Moi, Anthony and Cistac, Pierric and Rault, Tim and Louf, Remi and Funtowicz, Morgan and others},
  booktitle={Proceedings of the 2020 conference on empirical methods in natural language processing: system demonstrations},
  pages={38--45},
  year={2020}
}

@article{Kaplan2020ScalingLF,
  title={Scaling Laws for Neural Language Models},
  author={Jared Kaplan and Sam McCandlish and T. J. Henighan and Tom B. Brown and Benjamin Chess and Rewon Child and Scott Gray and Alec Radford and Jeff Wu and Dario Amodei},
  journal={ArXiv},
  year={2020},
  volume={abs/2001.08361},
  url={https://arxiv.org/pdf/2001.08361}
}

@inproceedings{dosovitskiy2021image,
  title={An Image is Worth 16x16 Words: Transformers for Image Recognition at Scale},
  author={Dosovitskiy, Alexey and Beyer, Lucas and Kolesnikov, Alexander and Weissenborn, Dirk and Zhai, Xiaohua and Unterthiner, Thomas and Dehghani, Mostafa and Minderer, Matthias and Heigold, Georg and Gelly, Sylvain and Uszkoreit, Jakob and Houlsby, Neil},
  booktitle={International Conference on Learning Representations},
  year={2021}
}

@inproceedings{radford2021learning,
  title={Learning Transferable Visual Models From Natural Language Supervision},
  author={Radford, Alec and Kim, Jong Wook and Hallacy, Chris and Ramesh, Aditya and Goh, Gabriel and Agarwal, Sandhini and Sastry, Girish and Askell, Amanda and Mishkin, Pamela and Clark, Jack and Krueger, Gretchen and Sutskever, Ilya},
  booktitle={Proceedings of the 38th International Conference on Machine Learning},
  series={Proceedings of Machine Learning Research},
  volume={139},
  pages={8748--8763},
  year={2021}
}

@misc{jaegle2021perceiver,
  title={Perceiver: General Perception with Iterative Attention},
  author={Jaegle, Andrew and Gimeno, Felix and Brock, Andrew and Zisserman, Andrew and Vinyals, Oriol and Carreira, Jo{\~a}o},
  year={2021},
  eprint={2103.03206},
  archivePrefix={arXiv},
  primaryClass={cs.CV}
}

@inproceedings{devlin2019bert,
  title={BERT: Pre-training of Deep Bidirectional Transformers for Language Understanding},
  author={Devlin, Jacob and Chang, Ming-Wei and Lee, Kenton and Toutanova, Kristina},
  booktitle={Proceedings of NAACL-HLT},
  pages={4171--4186},
  year={2019}
}

@article {Vishniakov2024.12.18.628606,
	author = {Vishniakov, Kirill and Viswanathan, Karthik and Medvedev, Aleksandr and Kanithi, Praveen K and Pimentel, Marco AF and Rajan, Ronnie and Khan, Shadab},
	title = {Genomic Foundationless Models: Pretraining Does Not Promise Performance},
	elocation-id = {2024.12.18.628606},
	year = {2024},
	doi = {10.1101/2024.12.18.628606},
	publisher = {Cold Spring Harbor Laboratory},
	URL = {https://www.biorxiv.org/content/early/2024/12/20/2024.12.18.628606},
	eprint = {https://www.biorxiv.org/content/early/2024/12/20/2024.12.18.628606.full.pdf},
	journal = {bioRxiv}
}

@article {ProteinGym,
	author = {Notin, Pascal and Kollasch, Aaron W. and Ritter, Daniel and van Niekerk, Lood and Paul, Steffanie and Spinner, Hansen and Rollins, Nathan and Shaw, Ada and Weitzman, Ruben and Frazer, Jonathan and Dias, Mafalda and Franceschi, Dinko and Orenbuch, Rose and Gal, Yarin and Marks, Debora S.},
	title = {ProteinGym: Large-Scale Benchmarks for Protein Design and Fitness Prediction},
	elocation-id = {2023.12.07.570727},
	year = {2023},
	doi = {10.1101/2023.12.07.570727},
	publisher = {Cold Spring Harbor Laboratory},
	journal = {bioRxiv}
}

@article{yang2018sixtyfive,
    title = {Sixty-five years of the long march in protein secondary structure prediction: The final stretch?},
    author = {Yang, Y. and Gao, J. and others},
    journal = {Briefings in Bioinformatics},
    volume = {19},
    number = {3},
    pages = {482--494},
    year = {2018}
}

@article{cuff1999evaluation,
    title = {Evaluation and improvement of multiple sequence methods for protein secondary structure prediction},
    author = {Cuff, J. A. and Barton, G. J.},
    journal = {Proteins: Structure, Function, and Bioinformatics},
    volume = {34},
    number = {4},
    pages = {508--519},
    year = {1999}
}

@article{abriata2018casp12,
    title = {Assessment of hard target modeling in CASP12 reveals an emerging role of alignment-based contact prediction methods},
    author = {Abriata, L. A. and Tamo, G. E. and others},
    journal = {Proteins: Structure, Function, and Bioinformatics},
    volume = {86},
    pages = {97--112},
    year = {2018}
}


\onecolumn
\appendix
\section{Appendix}

\subsection{Preliminary \textsc{ProtModernBERT} Experiments}
\label{app:preliminary}

\begin{figure*}[h!]
  \centering
  \includegraphics[width=0.95\textwidth]{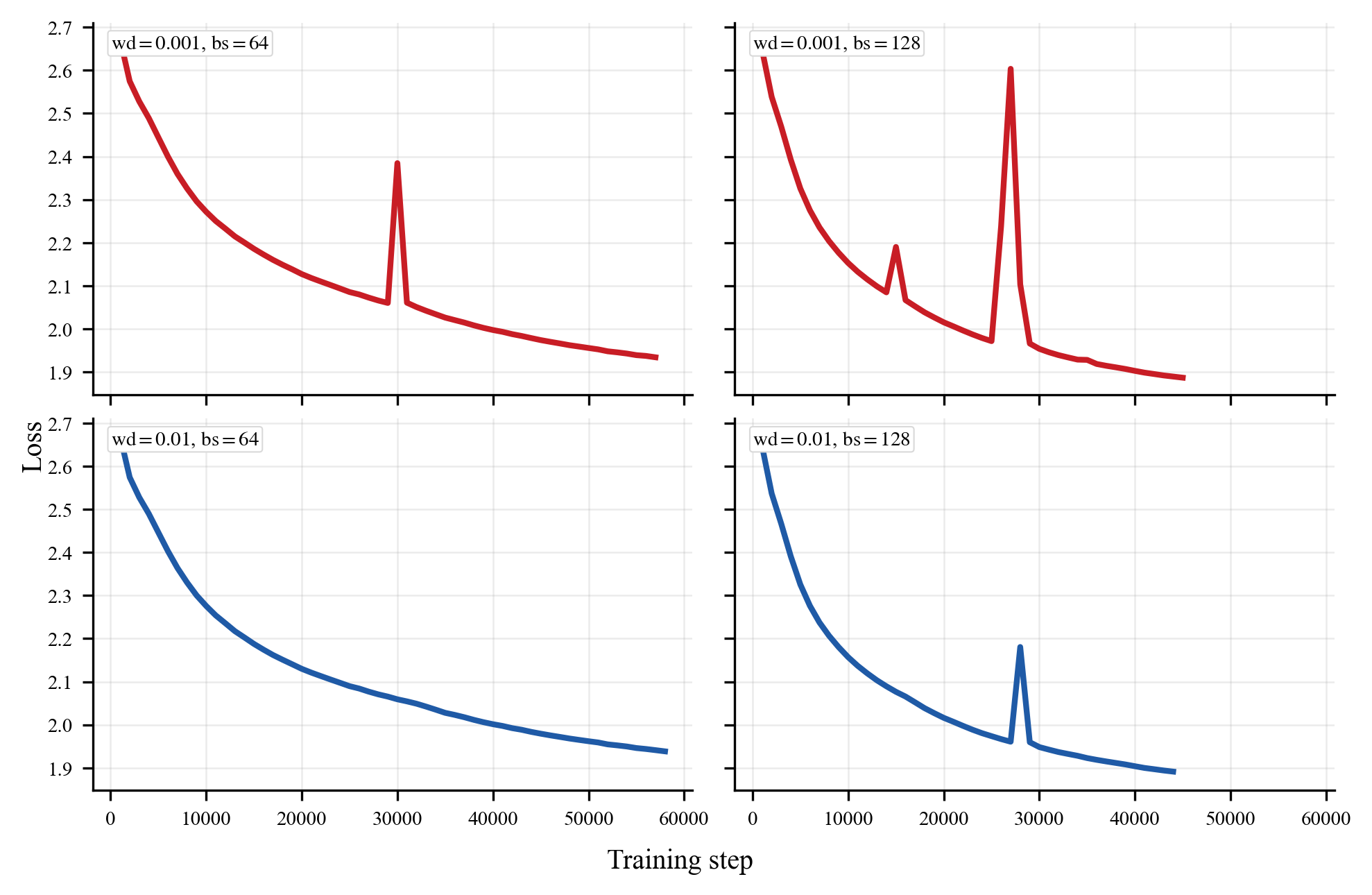}
  \caption{Training loss during the four preliminary pre-training runs varying weight decay (wd) and local batch size (bs). Lower weight decay  exhibits occasional loss spikes, while weight decay = 0.01 is more stable.}
  \label{fig:prelim-loss}
\end{figure*}

Before running the pre-training, we conducted 4 preliminary experiments. We used the same hyperparameters as described in Table \ref{tab:hyperparams}, except for local batch size and weight decay. We performed a grid search over weight decay values [0.01,0.001] and local batch sizes [64,128]. Each configuration was trained for 5 hours using 48 GPUs. We wanted to see how the training evolved with different weight decay values and batch sizes. 

In these experiments, we observed in Figure \ref{fig:prelim-loss} that with low weight decay values, there were steps where the model had loss spikes, while with 0.01, these steps were fewer and smaller. For training stability we chose 0.01.

Regarding the batch size, we selected 128 sequences per GPU, the largest value that our hardware could handle. One of the reasons is that in all the experiments we conducted, the training loss was greater than the evaluation loss, so we are far from overfitting. In other words, our model is generalizing well. Assuming that we are not overfitting, the experiment that made the most use of GPU resources was with a batch size of 128.



\subsection{Fine-tuning results}
\label{app:Fine-tuning results}

\begin{figure*}[h!]
\vskip 0.2in
\begin{center}
\centerline{\includegraphics[width=0.9\textwidth]{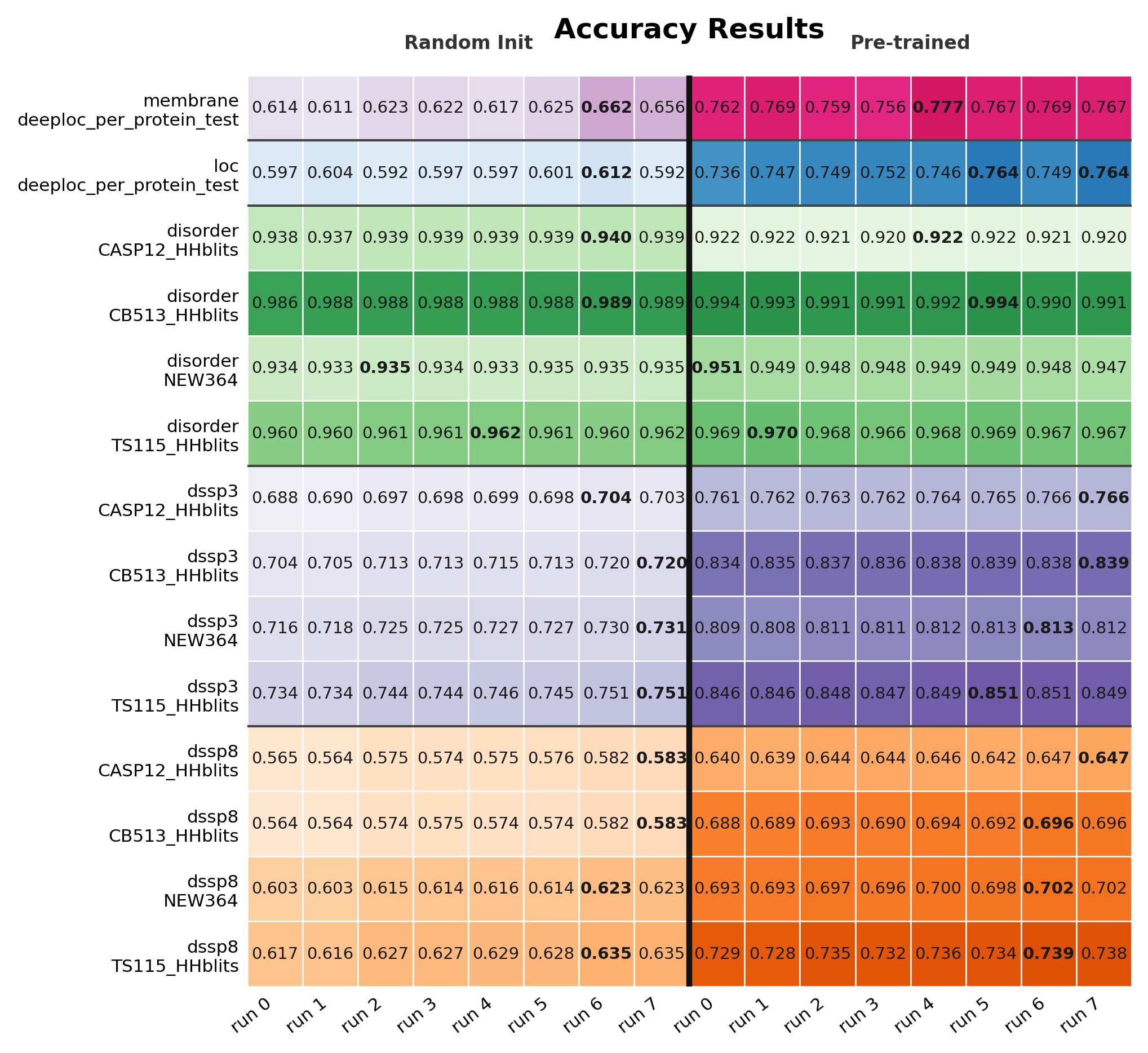}}
\caption{\textsc{ProtModernBERT} fine-tuning accuracy results.}
\label{fig:protmodernbert_finetuning_accuracy}
\end{center}
\vskip -0.2in
\end{figure*}

\begin{figure*}[h!]
\vskip 0.2in
\begin{center}
\centerline{\includegraphics[width=0.9\textwidth]{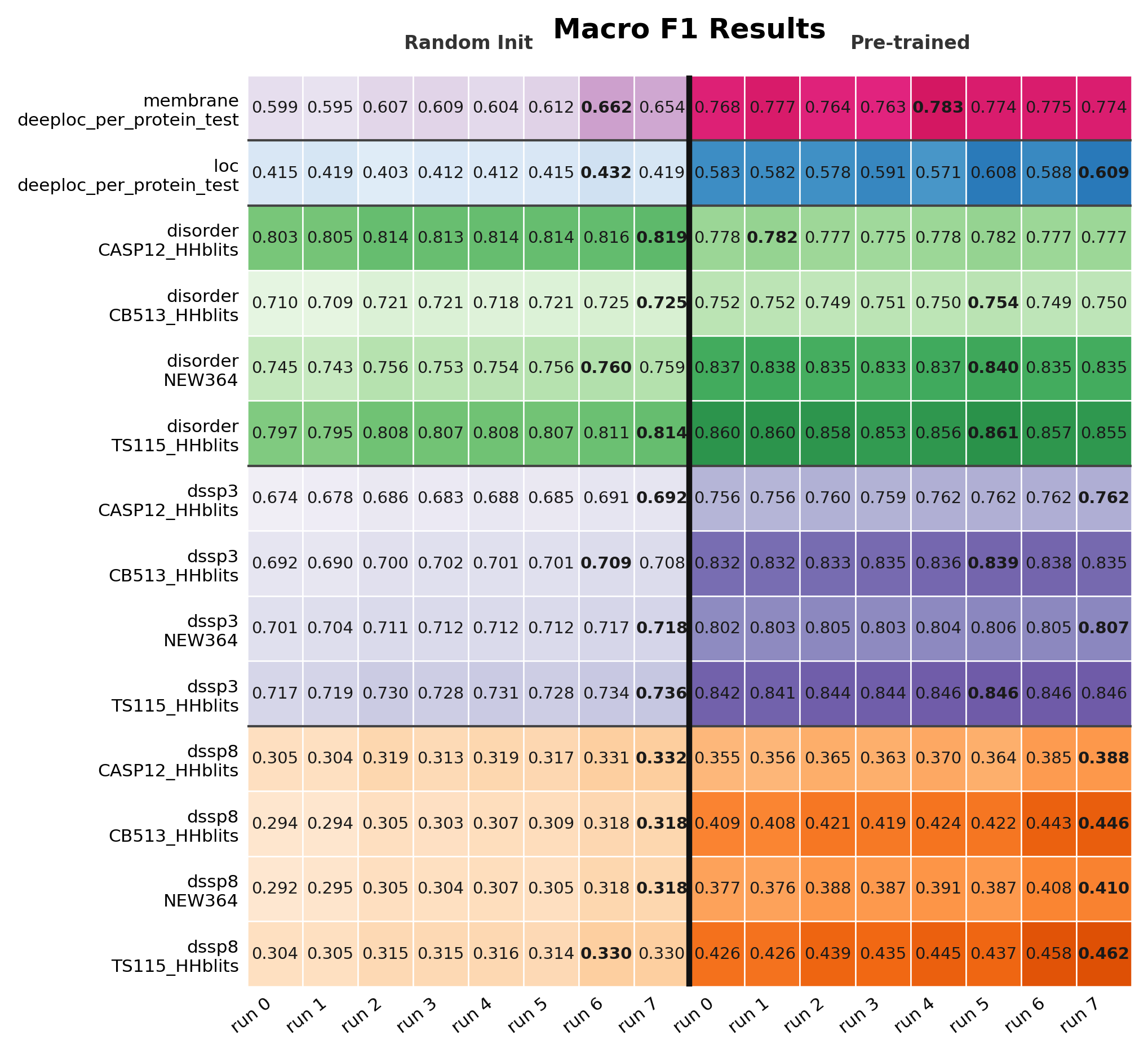}}
\caption{\textsc{ProtModernBERT} fine-tuning macro-F1 results.}
\label{fig:protmodernbert_finetuning_f1}
\end{center}
\vskip -0.2in
\end{figure*}

We fine-tune \textsc{ProtModernBERT} with different hyperparameters. We report the results on 8 different runs in Figures \ref{fig:protmodernbert_finetuning_accuracy} and \ref{fig:protmodernbert_finetuning_f1}. We present in Table \ref{tab:label_mapping_hyperparams_finetuning} a mapping between the experiment's label and its corresponding set of hyperparameters.

To ensure a rigorous and unbiased evaluation for the HHblits and DeepLoc benchmarks, we employ a validation protocol based on an internal partition of the training data. Specifically, we apply a cluster-based 80/20 split using MMseqs2, grouping sequences by identity to ensure that no close homologs appear in both the training and validation partitions. Hyperparameter tuning is performed over a grid including learning rates [3e-5, 5e-5], weight decay [0.02, 0.1], and an updated search range for training epochs. The configuration yielding the best performance on the validation partition is selected as the final setting. All benchmark results reported in the main text are obtained by evaluating these optimized configurations on the held-out test sets.
As shown in Table \ref{tab:best_hyperparameters}, this protocol confirms that our architectural inductive biases and pre-training gains remain robust across different validation splits. Performance deltas under this cluster-based partitioning remain stable, with gains of +0.02 to +0.08 observed on DSSP8 tasks under the updated hyperparameter grid.

Additionally, as described in Section~\ref{subsubsection:Fine-Tuning Experiments}, we present the final choice of hyperparameters not only for \textsc{ProtModernBERT} but for the rest of the models in Table \ref{tab:best_hyperparameters}. For each model, hyperparameters were selected based on downstream performance across all evaluated tasks. Specifically, for every candidate hyperparameter configuration explored during fine-tuning, we computed the average macro F1 score across all benchmarks. The configuration achieving the highest average macro F1 for a given model was selected as the final hyperparameter setting. This procedure was applied independently for each model and for both the pre-trained and random-initialization (\emph{randinit}) conditions. 

\begin{table}[h]
\centering
\small
\caption{Best hyperparameters used for each model when the model was fine-tuned under the cluster-based validation protocol.}
\label{tab:best_hyperparameters}
\begin{tabular}{lccc}
\hline
\textbf{Model} & \textbf{LR} & \textbf{Epochs} & \textbf{Weight Decay} \\
\hline
\textsc{Amplify} 120M  & 3e-5 & 4 & 0.02 \\
\textsc{Amplify} 120M RandInit & 3e-5 & 4 & 0.02 \\
\textsc{Amplify} 350M & 3e-5 & 4 & 0.02 \\
\textsc{Amplify} 350M RandInit & 3e-5 & 4 & 0.02 \\
\hline
ESM150M  & 5e-5 & 4 & 0.10 \\
ESM150M RandInit & 5e-5 & 4 & 0.10 \\
\hline
\textsc{ProtModernBERT}  & 5e-5 & 4 & 0.10 \\
\textsc{ProtModernBERT} RandInit & 5e-5 & 4 & 0.02 \\
Early \textsc{ProtModernBERT} & 5e-5 & 4 & 0.02 \\
\hline
\textsc{ProtBert}  & 3e-5 & 4 & 0.02 \\
\textsc{ProtBert} RandInit & 3e-5 & 4 & 0.02 \\
\hline
\end{tabular}
\end{table}



\begin{table}[h!]
\centering
\small
\caption{Mapping of labels and hyperparameters.}
\label{tab:label_mapping_hyperparams_finetuning}
\begin{tabular}{lccc}
\toprule
\textbf{Label} & \textbf{Learning Rate} & \textbf{Epochs} & \textbf{Weight Decay} \\
\midrule
Fine-tune 0 & 3e-5 & 4 & 0.02 \\
Fine-tune 1 & 3e-5 & 4 & 0.10 \\
Fine-tune 2 & 3e-5 & 6 & 0.02 \\
Fine-tune 3 & 3e-5 & 6 & 0.10 \\
Fine-tune 4 & 5e-5 & 4 & 0.02 \\
Fine-tune 5 & 5e-5 & 4 & 0.10 \\
Fine-tune 6 & 5e-5 & 6 & 0.02 \\
Fine-tune 7 & 5e-5 & 6 & 0.10 \\
\bottomrule
\end{tabular}
\end{table}

\subsection{Dummy majority-class baseline}
\label{app:dummy-baseline}

To contextualize the downstream results, we report a simple ``dummy'' baseline that predicts the most frequent label in the training set for every example in a given task. Concretely, for each dataset and task, we estimate the empirical label distribution on the training split and construct a classifier that always outputs the single most frequent label. We then evaluate this classifier on the corresponding test set using the same metrics as for our models, namely accuracy and macro-averaged F1.

This baseline captures the performance that can be achieved by exploiting only global class imbalance, without learning any sequence-dependent signal. As such, it provides a sanity check that the reported gains are not driven solely by skewed label distributions and yields a lower bound for meaningful performance on each dataset.

The dummy majority-class baseline results have been integrated into the left-most column of Figures ~\ref{fig:results_accuracy}  and ~\ref{fig:results_f1} for ease of comparison.

\subsection{\textsc{ESM2}-150M Training Compute Estimation}
\label{app:esm2-compute}

The original \textsc{ESM2} publications \cite{lin2023esm2} do not explicitly report training FLOPs for the 150M parameter variant, introducing uncertainty into direct compute comparisons. We provide two complementary estimates to bound the training cost.

\paragraph{Theoretical Lower Bound.}
We apply the standard $6NT$ approximation commonly used to estimate transformer training compute \cite{Kaplan2020ScalingLF}:
\begin{equation}
\text{FLOPs} \approx 6 \times N \times D
\end{equation}
where $N$ is the number of model parameters and $D$ is the total number of tokens processed during training. For \textsc{ESM2}-150M, we have $N = 1.5 \times 10^8$ parameters. According to the original paper, the model was trained for 500,000 steps with a batch size of 2,048,000 tokens per step, yielding:
\begin{equation}
D = 500,000 \times 2,048,000 = 1.024 \times 10^{12} \text{ tokens}
\end{equation}
Substituting into the $6NT$ formula:
\begin{equation}
\text{FLOPs} \approx 6 \times 1.5 \times 10^8 \times 1.024 \times 10^{12} \approx 9.2 \times 10^{20}
\end{equation}
This estimate reflects the core forward and backward pass computations but excludes several sources of overhead, including optimizer state updates, embedding and output layer operations, attention masking mechanics, data loading, inter-GPU communication, and padding inefficiencies. As a result, this value should be interpreted as a theoretical lower bound on actual training cost.

\paragraph{Empirical Upper Bound from \textsc{Amplify}.}
The \textsc{Amplify} paper \cite{Fournier2024.09.23.614603} provides data that allows us to derive an empirical upper bound for \textsc{ESM2}-150M training cost. The authors report that \textsc{Amplify}-120M required $1.6\times10^{21}$ FLOPs to train for 1 million steps, and note that \textsc{ESM2}-150M is more computationally expensive, requiring $1.25\times$ more FLOPs per training step. Thus, at 1 million steps, \textsc{ESM2}-150M would require approximately $2.0\times10^{21}$ FLOPs ($1.25 \times 1.6\times10^{21}$). Since we assume the \textsc{ESM2}-150M control was trained for only 500,000 steps (half the duration), as is the case in the original paper \cite{lin2023esm2}, the total compute is $1.0\times10^{21}$ FLOPs. This higher estimate likely accounts for practical overheads omitted in the $6NT$ approximation, providing a more realistic upper bound.

Based on these two approaches, we estimate that \textsc{ESM2}-150M was trained with a total compute budget in the range of $9\times10^{20}$--$1.0\times10^{21}$ FLOPs. This range places \textsc{ESM2}-150M's training cost below that of \textsc{Amplify}-120M ($1.6\times10^{21}$ FLOPs) and substantially below that of \textsc{ProtBert} ($6.0\times10^{21}$ FLOPs), underscoring the efficiency gains achieved through improved pre-training data curation and training recipes.

While \textsc{ESM2} typically achieves the strongest absolute performance, we calculate that the 150M parameter variant was trained on approximately 1.024 trillion tokens. This represents a data scale roughly 7.5 times larger than that of \textsc{ProtModernBERT}, which was pre-trained on 137B tokens. This observation is consistent with ESM2’s advantage being attributable, at least in part, to differences in pre-training scale and training recipe rather than architecture alone. This conclusion is further supported by our random-initialization (randinit) benchmarks, where the performance gap between \textsc{ESM2} and \textsc{ProtModernBERT} narrows considerably, demonstrating that their architectures exhibit comparable inductive biases when the benefit of extensive transfer learning is removed.

\end{document}